\PassOptionsToPackage{numbers,sort&compress}{natbib}
\documentclass[preprint,12pt]{elsarticle}

\usepackage{lineno}
\usepackage{comment}

\usepackage{amsmath,amssymb,amsfonts}
\usepackage{graphicx}
\usepackage{textcomp}
\usepackage[table]{xcolor}
\usepackage{subcaption}
\usepackage{hyperref}
\usepackage{balance}
\usepackage{dblfloatfix}
\usepackage[font={small,it}]{caption}
\usepackage{tabularx}
\usepackage{float}
\usepackage{svg}
\usepackage{makecell}
\usepackage{enumitem}
\usepackage{pifont}
\usepackage{multirow}
\usepackage{booktabs}
\usepackage{arydshln}
\usepackage{colortbl}
\usepackage{diagbox}
\usepackage[ruled,vlined]{algorithm2e}
\usepackage[cache=false]{minted}
\usepackage{microtype}
\journal{Information Sciences}

\begin{document}

\begin{frontmatter}
\author{Mohammad Meymani}
\ead{mohammad.meymani79@unb.ca}
\author{Roozbeh Razavi-Far}
\ead{roozbeh.razavi-far@unb.ca}
\affiliation{organization={Trustworthy and Secure AI (TSAI) Lab, Faculty of Computer Science, University of New Brunswick},
city={Fredericton}, country={Canada}}

\title{{Divided We Fall: Defending Against Adversarial Attacks via Soft-Gated Fractional Mixture-of-Experts with Randomized Adversarial Training}}

\begin{abstract}
{Machine learning is a powerful tool enabling full automation of a huge number of tasks without explicit programming. Despite recent progress of machine learning in different domains, these models have shown vulnerabilities when they are exposed to adversarial threats. Adversarial threats aim to hinder the machine learning models from satisfying their objectives. They can create adversarial perturbations, which are imperceptible to humans' eyes but have the ability to cause misclassification during inference. In this paper, we propose a defense system, which devises an adversarial training module within mixture-of-experts architecture to enhance its robustness against white-box evasion attacks. In our proposed defense system, we use nine pre-trained classifiers (experts) with ResNet-18 as their backbone. During end-to-end training, the parameters of all experts and the gating mechanism are jointly updated allowing further optimization of the experts. Our proposed defense system outperforms prior MoE-based defenses under strong white-box FGSM and PGD evaluation on CIFAR-10 and SVHN. The use of multiple experts increases training time and compute relative to single-network baselines; however, inference scales approximately linearly with the number of experts and is substantially cheaper than training}.
\end{abstract}
\end{frontmatter}
\section{Introduction}
{Machine learning (ML) is a cornerstone of artificial intelligence and is widely deployed in applications such as computer vision \cite{khan2020machine}, healthcare \cite{qayyum2020secure}, natural language processing \cite{omar2022robust}, cybersecurity \cite{apruzzese2023role}, and autonomous systems. Despite its success, the widespread adoption of ML has exposed inherent vulnerabilities that adversaries can exploit through adversarial attacks, threatening the integrity, privacy, and availability of deployed systems \cite{pitropakis2019taxonomy}. Adversarial machine learning (AML) lies at the intersection of cybersecurity and ML, studying attacks that manipulate model behavior and developing defenses to mitigate them. Depending on the threat model, adversaries may possess partial or full knowledge of the model and training data, enabling the generation of adversarial examples that induce misclassification at inference time \cite{bountakas2023defense}. Figure \ref{fig:adversarial effect} shows how an adversarial sample looks like on an MNIST \cite{lecun2002gradient} sample when the adversary has a perfect knowledge about the model's parameters. Samples like this can cause model's misclassification with a high confidence during test time.

\begin{figure}[h]
    \centering
    \includegraphics[width=0.75\linewidth]{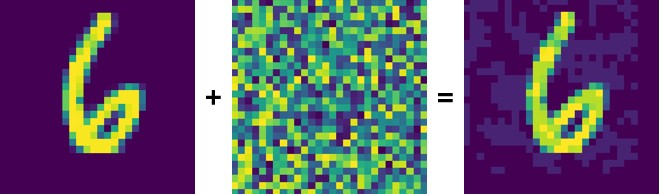}
    \caption{Generating an adversarial example on an MNIST digit by adding a carefully crafted perturbation, which remains imperceptible to humans but successfully fools the target models.}
    \label{fig:adversarial effect}
\end{figure}

These risks have elevated adversarial ML to a global concern. Government and regulatory bodies have identified adversarial and generative AI as threats to privacy, security, and public safety, emphasizing robustness, reliability, and transparency as key requirements for trustworthy AI systems \cite{AI_Task_Force_2024}. Such concerns underscore the urgent need for robust ML models that remain reliable under adversarial conditions. Robust machine learning, a core component of trustworthy AI, focuses on designing defense mechanisms that mitigate adversarial threats while preserving performance in benign settings \cite{kaur2022trustworthy}. Among emerging approaches, mixture-of-experts (MoE) architectures offer a promising direction by enabling specialization and adaptive routing, which can enhance resilience to adversarial perturbations while maintaining efficiency.}

{We consider a white-box adversarial setting, where the attacker has full access to the model parameters, gradients, architecture, and training procedure. The attacker’s objective is to induce misclassification at inference time by crafting adversarial examples under an $\ell_\infty$ norm. The attacker has white-box knowledge and optimizes perturbations accordingly. We do not consider data poisoning, model extraction, or privacy attacks in this work; our evaluation focuses exclusively on evasion robustness.}

In pursuit of building a secure system, we build a novel defense system named \underline{D}ivided \underline{W}e \underline{F}all (DWF). In DWF, we employ \underline{M}ixture \underline{o}f \underline{E}xperts (MoE) architecture which consists of benign and adversarial experts. The training scheme in DWF is joint, meaning that we do not freeze the weights of our pretrained experts during end-to-end training, which leads to stronger generalization capability of our proposed defense system improving the clean accuracy and robustness of our model. {To ensure rigorous evaluation, robustness is assessed using strong white-box attacks, including properly tuned multi-step PGD with many iterations, multi-step restarts, and adaptive variants that optimize through the full mixture-of-experts architecture.}

{The proposed method is termed Divided We Fall to emphasize its distributed robustness principle. Rather than relying on a single monolithic classifier, which can be entirely compromised by adversarial perturbations, DWF decomposes the model into multiple specialized experts trained on different regimes (benign, FGSM, and PGD). The soft gating mechanism adaptively combines their outputs at inference time, allowing the system to remain reliable even if individual experts are partially affected. By avoiding a single point of failure and encouraging specialization and cooperation, the overall model becomes more resilient to adversarial attacks.}

{Our contributions are as follows:\\
Methodological contributions:
\begin{itemize}
    \item We introduce a new training approach for MoE, where pretrained experts are jointly updated along with the gating mechanism during an end-to-end training session, where we use a mixed batch encompassing both benign and adversarial data.
    \item The randomized attack settings during all training stages enable DWF to generalize more efficiently in comparison to its competitors.
\end{itemize}
Empirical findings:
\begin{itemize}
    \item Under a white-box $\ell_\infty$ threat model with FGSM and multi-step PGD attacks, DWF achieves higher robust accuracy than prior MoE-based defenses (ADVMoE, SoE) on CIFAR-10 and SVHN.
    \item DWF preserves high clean accuracy (>90\%) while improving robustness relative to the defended baselines.
    \item Robustness improvements remain consistent under stronger attacks, including multi-step and adaptive PGD, demonstrating stable performance across evaluation protocols.
\end{itemize}}

In Section \ref{sec:backgorund-problem-setting}, we illustrate the background problem setting. In Section \ref{sec:related-works}, we introduce similar defense systems. In Section \ref{sec:methodology}, we present the methodology and workflow of our defense. In Section \ref{sec:experiments}, we demonstrate our experimental results. {In Section \ref{sec:limitations-future-works}, we investigate our proposed defense system limitations and potential future directions. Finally, in Section \ref{sec:conclusion}, conclude our work.}
\section{Background and Problem Setting}
\label{sec:backgorund-problem-setting}
{In this section, we introduce MoE architecture and a background on adversarial threat model.}
\subsection{Mixture-of-Experts}
\begin{figure}[h]
    \centering
    \includegraphics[width=0.75\linewidth]{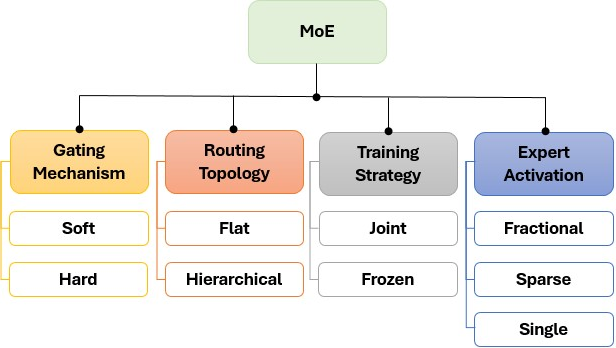}
    \caption{Main components and types of mixture of experts.}
    \label{fig:moe-taxonomy}
\end{figure}

MoE is an ensemble architecture using multiple experts, a gating mechanism to assign weights, and a router to aggregate results \cite{mu2025comprehensive}. MoEs can be categorized based on their gating mechanism, routing topology, training strategy, and experts activation, which is shown in Figure \ref{fig:moe-taxonomy}.

Based on gating mechanism, MoE could be either soft or hard; in soft MoE the gating assigns non-zero weights to each expert forming fractional activation of experts, while in hard MoE, the gating mechanism chooses a subset of experts and assign wights to them making experts activation as sparse, or single if only one of them is chosen \cite{puigcerver2023sparse}.

Training strategy could be either joint or frozen. In the first one, both router and experts are trained simultaneously, while the latter one uses pretrained experts and freezes their weights during training of the router \cite{mu2025comprehensive}.

Based on the routing topology we have flat and hierarchical routing schemes. In the flat MoE, all the experts reside in a single layer and the router chooses from them. In the hierarchical topology, the first router groups the experts into subgroups, and, then, the second one chooses some experts from these subgroups \cite{mu2025comprehensive}.

{Using an expert-driven model allows for specialization within each part of the system, much like Unmanned Aerial Vehicle (UAVs) assigned specific roles within a swarm. These experts could respond dynamically to different types of attacks or threats, providing targeted defenses that are optimized for specific conditions, rather than relying on a single, monolithic defense mechanism \cite{bakirci2023novel}.}

{Our design adopts soft routing with fractional expert activation in a flat MoE architecture to balance robustness, stability, and computational cost. Soft gating enables smooth expert weighting, avoiding brittle decision boundaries associated with hard routing and improving gradient flow during adversarial training. Fractional activation allows multiple experts to contribute to each prediction, increasing functional diversity and reducing single-expert failure under adversarial perturbations, at the cost of moderately increased computation compared to sparse activation. We choose a flat architecture to maintain simplicity and reproducibility, while still enabling specialization through independent experts. Overall, these choices prioritize robustness and graceful degradation under attack, while keeping the computational overhead manageable.}
\subsection{General Adversarial Threat Models}
Adversarial threats can be viewed from different angles: attacker's knowledge, type, specificity, and objective.
\subsubsection{Attacker's Knowledge}
Attacker's knowledge shows the access level of the adversary to model's parameters, gradients, architecture, and training data. This knowledge is categorized into white-box, black-box, and gray-box based on the adversary's access level \cite{pitropakis2019taxonomy}.
\paragraph{White Box}
Adversaries with white-box knowledge are the most effective and destructive attackers. These attackers have a full access to model's architecture, parameters, and gradients during training with full knowledge on the training data; this enables the attackers to generate adversarial samples based on the gradients and training data, which makes them very powerful. However, in a real-world scenario, adversaries can't achieve such an access, making them more suitable for research and testing the models' robustness \cite{pitropakis2019taxonomy}.
\paragraph{Black Box}
In black-box scenario, unlike white-box, the adversaries have no knowledge about the model and the presence of any defense system. The adversaries can only give inputs and receive the corresponding outputs \cite{lin2023sensitive}.
\paragraph{Gray Box}
Gray box knowledge lies between white box knowledge and black box knowledge. The adversary has some degree of knowledge about the target system, which is not as perfect as white box, and is not as scarce as black box knowledge \cite{pitropakis2019taxonomy}.
\subsubsection{Attacker's specificity}
Based on specificity, each attack could either be targeted or untargeted. In targeted attacks, the adversary aims to degrade the model's performance when it faces specific patterns/inputs, while in untargeted attacks, also named to as indiscriminate, the goal is to cause any misclassification \cite{bountakas2023defense}.
\subsubsection{Attacker's objective}
Based on the objective, the adversary can follow degrade, breach, or both scenarios. In the first scenario, the adversary aims to harm the system's performance and/or violets its integrity. In breach, the adversary aims to explore the model in order to steal sensitive information, without affecting the model's performance \cite{pitropakis2019taxonomy}.
\subsubsection{Attacker's type}
\begin{figure}[h]
    \centering
    \includegraphics[width=0.75\linewidth]{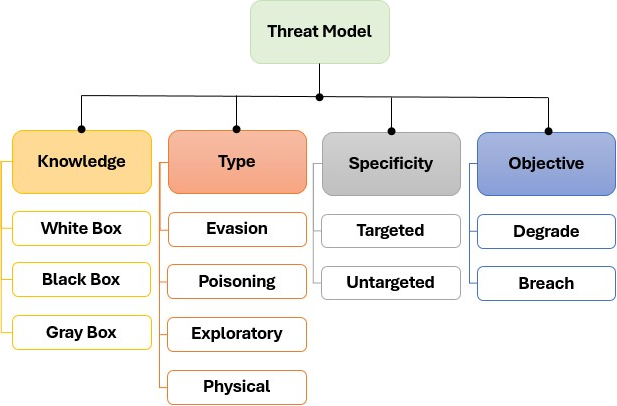}
    \caption{General taxonomy of the threat model.}
    \label{fig:threat-model}
\end{figure}

In a general perspective, an adversarial attack belongs to one of the following types: evasion, poisoning, backdoor, or physical layer attacks.
\paragraph{Evasion}
An evasion attack targets the model at test time without affecting the system's integrity on clean data. Evasion attacks are the most practical attack types aiming to degrade the model's performance by causing misclassification during testing \cite{pitropakis2019taxonomy}. For instance attacks such as Fast Gradient Sign Method (FGSM) \cite{goodfellow2014explaining}, Projected Gradient Descent (PGD) \cite{madry2017towards}, and  Carlini \& Wagner (C\&W) \cite{carlini2017towards}  make use of the model's gradient during training to generate samples which evade the classifier during the test time. Equations \ref{eq:fgsm-formula} and \ref{eq:PGD-Formula}, in Section \ref{sec:experiments}, show how an adversarial sample is generated in FGSM and PGD, respectively.
\paragraph{Other Attack Types}
{Poisoning attacks compromise a model's integrity by corrupting training data or embeddings, either through label manipulation or clean label attacks \cite{tian2022comprehensive}. Exploratory attacks aim to violate privacy of the system. These attacks probe the target model to gain information about model's parameters, training data, and algorithms on without affecting the performance and integrity of the model \cite{zhang2024defending}. Moreover, physical attacks are launched on the models, which are deployed on a special physical hardware such as FPGA, GPU, and edge devices to exploit vulnerabilities emerged due to the increased attack surface in the physical layer \cite{regazzoni2020machine}. In this research work, we only focus on defending against evasion attacks to assess the resilience of our proposed approach under severe white-box attacks.} Figure \ref{fig:threat-model} shows an overview of adversarial threat model.

{We evaluate inference-time robustness under a white-box $\ell_{\infty}$ threat model, where the adversary constructs an adversarial example $x_{\text{adv}} = x + \delta$ subject to $\|\delta\|_{\infty} \le \epsilon$, with the goal of inducing misclassification. In our experiments, we report results under FGSM (single-step) and PGD (multi-step) attacks (including Multi-Step PGD and Adaptive PGD) using the settings summarized in Table \ref{tab:attack-settings}.  After each update step, the perturbation is projected back into the $\ell_{\infty}$ ball around the clean input, $x - \epsilon \le x_{\text{adv}} \le x + \epsilon$, and valid pixel bounds are enforced by clipping  the perturbed sample to the input range $[0,1]$ using  $x_{\text{adv}} = \texttt{clamp}(x_{\text{adv}}, 0, 1)$. In this work, "imperceptible" perturbations are defined operationally by the chosen  $\ell_{\infty}$ budgets ($\epsilon \in \{0.01, 0.02, \ldots, 0.09\}$ for FGSM and  $\epsilon = \frac{8}{255}$ for PGD-based attacks), which correspond to standard small-magnitude perturbations used in robustness evaluations.}

All together, what we mentioned illustrates the seriousness of adversarial threats, highlighting the urge to build robust models to defend these adversarial threats, emphasizing the importance of adversarial machine learning and robust models.
\section{Related Works}
\label{sec:related-works}
Given the arms race nature of adversarial machine learning, a plethora of attacks and defenses have been proposed and evolved over the years. This highlights the evolving nature of AML. In the following paragraphs, we investigate the related defenses in this area.

Adversarial training is a defense strategy aiming to train the model on both benign and adversarial samples to increase generalization capability of the model enabling it to correctly predict both benign and adversarial inputs \cite{bai2021recent}. However, this method often comes with a trade-off \cite{zhang2025catil}, where the model sacrifices it accuracy on the benign data to increase its robustness against adversarial inputs \cite{zhao2024adversarial}.

Adversarial training was first introduced in \cite{goodfellow2014explaining}, which increased the model's robustness against FGSM attacks by sacrificing the clean accuracy of the model. Methods such as TRadeoff-inspired Adversarial
DEfense via Surrogate-loss minimization (TRADES) \cite{zhang2019theoretically} and Fair Robust Learning (FRL) \cite{xu2021robust} have been introduced to control natural robustness-accuracy trade-off in adversarial training. Additionally, methods such as Fast Free Adversarial Training (FFAT) \cite{shafahi2019adversarial} tackles the computational overhead of adversarial sample generation, while having a lower accuracy and robustness in comparison to standard adversarial training.

Ensemble architectures have shown strong robustness against adversarial threats in comparison to the non-ensemble architectures. This is due to the natural complex architecture of ensemble models. Ensemble-based defense methods are categorized into distillation, hierarchical, and MoE.

Ensemble distillation-based defenses aim to harden the model by training it on soft labels and/or noisy logits instead of hard labeling, producing results via ensembles. Research works in this category include Distillation Hardening for Random Forest (DDH-RF) \cite{apruzzese2020hardening} and SELENA \cite{tang2022mitigating}. Hierarchical defenses, on the other hand, aim to build ensemble of models in a hierarchical manner to detect and mitigate adversarial attacks in multiple levels, which make the defense systems robust against a wider range of attacks. Hierarchical defenses include Application Constraints (AppCon) \cite{apruzzese2020appcon}, Model Stacking \cite{salem2018ml}, and Ensemble Methods as a Defense (EMD) \cite{strauss2017ensemble}.

\cite{puigcerver2022adversarial} {and} \cite{pavlitska2024towards} investigate the adversarial robustness of MoEs. In \cite{han2024enhancing} {and} \cite{zhang2025optimizing}, some defense systems are proposed, which aim to protect the architecture of MoEs. Defenses such as adversarial mixture-of-expert (ADVMoE) \cite{zhang2023robust} and synergy-of-experts (SoE) \cite{cui2022synergy} make use of the ensemble architecture of MoEs to defend against adversarial threats. Both ADVMoE and SoE use adversarial training techniques, while training their MoE to increase the robustness of their model. 

{DWF is not a direct recombination of MoE and standard adversarial training as used in prior MoE defenses (e.g., SoE, ADVMoE). Concretely, DWF differs along the three axes the MoE pipeline exposes: training pipeline, routing, and activation. Unlike SoE/ADVMoE, which use hard routing with single/sparse expert activation under standard adversarial training, DWF adopts soft gating with fractional expert activation (weighted combination of all experts) and trains the system using a pretrain-then-joint scheme with randomized adversarial training during training. Table \ref{tab:moe-architecture-comparison} presents the differences between our proposed approach and other competitors.}

\begin{table*}[h]
    \centering
    \caption{MoE type of DWF vs. other competitors - gating mechanism (GM), routing topology (RT), training strategy (TS), experts activation (EA), and {adversarial training (AT)}.}
    \begin{tabular}{l c c c c c}
        \hline
        Defense &  GM & RT & TS & EA & AT type\\
        \hline
        SoE&Hard&Flat&Joint&Single&Standard\\
        ADVMoE&Hard&Flat&Joint&Sparse&Standard\\
        
        DWF&Soft&Flat&Pre-trained+Joint&Fractional&Randomized\\
        \hline
    \end{tabular}
    \label{tab:moe-architecture-comparison}
\end{table*}

{While standard adversarial training methods improve robustness by augmenting training data with adversarial examples, they typically train a single model with a uniform objective. In contrast, our approach explicitly decomposes the problem into multiple specialized experts, each trained on different regimes (benign, FGSM, and PGD), and combines them via a soft gating mechanism. This distinguishes totally from conventional adversarial training, which lacks explicit expert specialization and input-dependent routing.}

{While mixture-of-experts defenses are the primary architectural focus of this work, it is important to position DWF relative to widely used non-MoE robustness baselines. Standard adversarial training methods improve robustness by optimizing a single model against worst-case perturbations. These approaches represent strong and widely adopted robustness baselines, but they lack explicit expert specialization and input-dependent routing.

Our evaluation includes adversarially trained ResNet-18 models (FGSM-AT and PGD-AT) as representative non-MoE baselines. The results show that DWF achieves higher robust accuracy while preserving stronger clean accuracy compared to these single-network defenses. The primary comparison with MoE-based methods (SoE and ADVMoE) is intentional, as it enables controlled architectural comparisons that isolate the impact of soft gating, fractional expert activation, and randomized adversarial training within the MoE paradigm.}

{To ensure that the observed robustness does not stem from gradient masking or obfuscated gradients, all attacks are implemented in a true white-box setting and differentiate through the full mixture-of-experts architecture, including all experts, the gating network, and the final loss. Perturbations are optimized end-to-end with respect to the complete model, preventing gradient obfuscation through routing or expert specialization. Robust accuracy decreases monotonically as attack strength increases (larger $\epsilon$ and more PGD iterations), which is inconsistent with gradient masking behavior. Moreover, inference in DWF is deterministic; therefore, expectation over transformation (EoT) with multiple samples yields identical gradients and attack outcomes, confirming the absence of stochastic masking.}
\section{Methodology}
\label{sec:methodology}
\begin{figure*}[h]
    \centering
    \includegraphics[width=\linewidth]{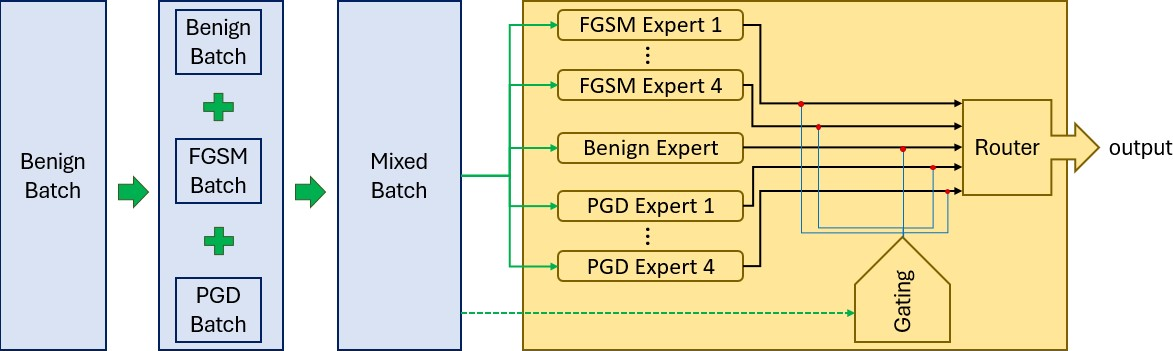}
    \caption{General view of the end to end training of our proposed defense system.}
    \label{fig:workflow}
\end{figure*}

Our defense uses soft and flat structure for MoE with fractional activation of experts. {In the first phase of training, we independently train a classifier on benign data (benign expert), four classifiers on FGSM data (FGSM experts), and four classifiers on PGD data (PGD experts). This design encourages diversity among experts, which is a key principle in mixture-of-experts architectures.} After this pretraining stage, we move to end-to-end learning, in which the entire framework is jointly optimized. During this stage, for each batch of data, we generate adversarial variants of the inputs using both FGSM and PGD attacks, and augment them with the original benign inputs to form an expanded batch. This mixed batch preserves the diversity of benign and adversarial samples, enhancing the network’s robustness and generalization capability. Finally, the combined batch is fed into the network, so the whole framework learns both benign and adversarial patterns. Figure \ref{fig:workflow} shows an overview of our defense model's workflow.

{During training, we sample attack strengths (FGSM $\epsilon$ and PGD iterations) from a range rather than fixing a single configuration. This strategy prevents overfitting to a specific perturbation magnitude and exposes the model to adversarial examples of varying difficulty, improving robustness generalization across threat levels. Consequently, the model is not tuned to a particular evaluation $\epsilon$ but learns to remain stable under a spectrum of perturbations. A detailed ablation over all sampling ranges is left for future investigation due to computational cost.}

To construct FGSM experts, we need to pre-train them before end-to-end training. During adversarial training of an FGSM expert, within every batch and each epoch we randomly assign a value between 0.01 to 0.09 to perturbation size  $\epsilon$, and the models is adversarially trained on random perturbation size during that epoch. This strategy helps to increase generalization ability of the network, making it more robust against different perturbation sizes. We train four models in this manner. Algorithm \ref{algorithm:FGSM-training} shows how we train the FGSM experts.

\begin{algorithm}[h]
\caption{Training FGSM Experts.}
\KwIn{Benign inputs: $X$, labels: $Y$, and number of epochs: $E$.}
model $\gets$ Instantiate(ResNet-18)\;
\For{$e \in \{0,~1,~\dots,~E-1\}$}{
  \For{$(x,~y) \in (X,~Y)$}{
    $\epsilon \gets \textbf{Random}\{0.01,~0.02,~\dots,~ 0.09\}$\;
    $x_{\text{adv}} \gets x + \epsilon \cdot \text{sign}(\nabla_x L(\theta,~x,~y))$\;
    $y_p \gets \text{model.forward}(x_{\text{adv}})$\;
    loss $\gets \ell(y_p,~y)$\;
    model.backward(loss)\;
  }
}
\Return model.\
\label{algorithm:FGSM-training}
\end{algorithm}

We take the same strategy to train the PGD experts, with a minor difference. This time we alter the iterations number, and randomly choose a different iteration number within in each batch. With the same approach we train four PGD experts. Algorithm \ref{algorithm:PGD-training} shows how we train the PGD experts.

\begin{algorithm}[h]
\caption{Training PGD Experts.}
\KwIn{Benign images: $X$, labels: $Y$, perturbation bound: $\epsilon$, step size: $\alpha$, and number of epochs: $E$.}
model $\gets$ Instantiate(ResNet-18)\;
\For{$e \in \{0,~1,~\dots,~E-1\}$}{
  \For{$(x,~y) \in (X,~Y)$}{
    iterations $\gets \textbf{Random}\{10,~20,~30,~40,~50\}$\;
    $x_{\text{adv}} \gets \text{PGD}(x,~\epsilon,~\alpha,~ \text{iterations})$\;
    $y_p \gets \text{model.forward}(x_{\text{adv}})$\;
    loss $\gets \ell(y_p,~y)$\;
    model.backward(loss)\;
  }
}
\Return model.\
\label{algorithm:PGD-training}
\end{algorithm}

Now we have a benign expert, which is trained on benign data, four FGSM experts, and four PGD experts. To train the end-to-end framework, we do not freeze the parameters of the experts. During each batch we generate FGSM and PGD version of our input, with random perturbation size and iteration number, respectively. The batch size during this stage of training is three times larger than the batch size for training the experts, which includes benign input, FGSM inputs, and PGD inputs that together can enhance clean accuracy and robustness of the network. Algorithm \ref{algorithm:End2End-training} shows the end-to-end training of our MoE network.

\begin{algorithm}[h]
\caption{End-to-End Training of Mixture of Experts.}
\KwIn{Benign images: $X$, labels: $Y$, perturbation bound: $\epsilon$, step size: $\alpha$, number of epochs: $E$, and Experts: experts.}
model $\gets$ Instantiate(mixture-of-expert)\;
model.add(experts)\;
\For{$e \in \{0,~1,~\dots,~E-1\}$}{
  \For{$(x, y) \in (X, Y)$}{
    $\epsilon_\text{FGSM} \gets \textbf{Random}\{0.01, 0.02, \dots, 0.09\}$\;
    $x_\text{FGSM} \gets x + \epsilon_\text{FGSM} \cdot \text{sign}(\nabla_x L(\theta, x, y))$\;
    iterations $\gets \textbf{Random}\{10, 20, 30, 40, 50\}$\;
    $x_\text{PGD} \gets \text{PGD}(x,\epsilon, \alpha, \text{iterations})$\;
    $x_\text{mixed} = \text{Concatenate}(x, x_\text{FGSM}, x_\text{PGD})$\;
    $y_p \gets \text{model.forward}(x_\text{mixed})$\;
    $y_\text{mixed} = \text{Concatenate}(y,y,y)$\;
    loss $\gets \ell(y_p, y_\text{mixed})$\;
    model.backward(loss)\;
  }
}
\Return model.\
\label{algorithm:End2End-training}
\end{algorithm}

{During adversarial training, adversarial examples are generated with respect to the full end-to-end loss of the mixture-of-experts model, allowing gradients to propagate through all experts and the gating mechanism. No random restarts are used during training. For FGSM, a single-step gradient update is applied using a randomly sampled perturbation magnitude. For PGD, iterative updates are performed with projection onto the $\ell_\infty$ ball around the clean input after each step. In all cases, perturbed samples are clipped to the valid input range $[0,1]$ to ensure valid pixel values.}

After all these training stages, the model is ready for testing and deployment. During both phases, DWF uses fractional schema for the experts, which enables them to contribute in the final decision. Final decision is made as follows:
\begin{equation}
\label{eq:moe}
y = \sum_{i =1}^{n}w_i(x) \cdot f_i(x)
\end{equation}
where $n$ is the number of experts, $w_i(x)$ is the gating weights, and $f_i(x)$ stands for the $i_{th}$ expert.

{The gating function is implemented as a lightweight learnable router that takes the same input $x$ as the experts and produces a soft assignment vector using a small multilayer perceptron followed by a softmax normalization. The router first flattens the input image and then applies four fully connected layers with sizes 1024, 512, 128, and the number of experts, respectively. ReLU activation is used after each hidden layer, and the final Softmax layer produces non-negative routing weights that sum to one. The router is trained jointly with the experts and does not share parameters with the expert backbones.}

The final prediction is obtained via a weighted sum of all expert logits (soft routing), rather than hard or sparse selection. All experts and the gating network are trained jointly in an end-to-end manner using the standard cross-entropy loss. No auxiliary routing objectives, load-balancing losses, temperature scaling, or sparsity regularization are used. This simple design avoids additional hyperparameters and improves reproducibility.

{The weights $w_i(x)$ in (\ref{eq:moe}) are generated by a learnable gating network $g(.)$. Given an input $x$, the gating network produces a score vector $s(x) \in \mathbb{R}^n$, which is normalized using a softmax function to obtain non-negative weights that sum to one:
\begin{equation}
    w_i(x) = \frac{e^{s_i(x)}}{\sum_{j=1}^ne^{s_j(x)}}
\end{equation}

The gating network is trained jointly with all experts during end-to-end training by backpropagating the cross-entropy loss through both the experts and the gating mechanism. Thus, $w(x)$ is not fixed or heuristic, but learned from data in an input-dependent manner.}

{During training, we randomize attack parameters such as perturbation magnitude and step size. This randomness is not strictly required for robustness, but is used to increase perturbation diversity and avoid overfitting to a single fixed attack configuration. By exposing the experts to a wider range of adversarial strengths, the model generalizes better to unseen attacks. In practice, similar behavior can still be achieved with fixed parameters, although randomized training typically provides slightly more stable robustness.}

{The number of experts determines a trade-off between robustness and computational cost. A larger number of experts enables greater specialization to different input regimes, which can improve robustness. However, both memory and inference cost grow approximately linearly with the number of experts, while the marginal gain diminishes as experts begin to learn overlapping representations. Therefore, we adopt nine experts as a practical compromise that provides strong robustness while maintaining reasonable computational efficiency.}

{All experts share the same backbone architecture (ResNet-18) in our implementation. This choice is intentional to ensure fair comparisons and to isolate the contribution of the proposed soft-gated mixture-of-experts training strategy, rather than improvements arising from heterogeneous model capacities. Despite identical architectures, diversity is still achieved through specialization: each expert is trained on different input regimes (benign and adversarial attacks), which results in distinct decision boundaries and complementary behaviors. Incorporating heterogeneous backbones could further increase diversity and potentially improve robustness; however, this would introduce additional computational cost and confounding factors. Exploring architectural diversity among experts is an interesting direction for future work.}

{Randomization is applied only during training, where adversarial parameters are sampled from predefined ranges (e.g., $\epsilon$ for FGSM and iteration counts for PGD) to expose the model to diverse perturbation strengths. This improves robustness generalization. During inference, no randomness is used: the gating mechanism and experts operate deterministically, and the model produces a fixed output for a given input. Therefore, expectation over
transformation (EoT) is not required at test time, and using EoT samples $>1$ yields the same gradients and attack
outcomes; thus we set EoT samples $=1$ in the adaptive evaluation.}

{Mixture-of-experts models may suffer from expert collapse, where routing concentrates on a small subset of experts. Our design mitigates this risk through architectural choices rather than explicit regularization. We employ soft gating with fractional activation, which assigns a normalized weight to every expert for each input, ensuring that predictions are formed through a weighted combination rather than exclusive selection. This contrasts with hard or sparse routing, where only a few experts are active and collapse is more likely. Furthermore, experts are pretrained on different input regimes (benign, FGSM, and PGD), encouraging functional diversity and reducing the likelihood that a single expert dominates across all inputs. While we do not incorporate load-balancing losses or diversity regularizers, these design choices provide a theoretical basis for promoting balanced expert utilization.}

{Fractional activation evaluates all experts using continuous routing weights, resulting in a theoretical increase in inference cost that scales approximately linearly with the number of experts relative to a single backbone of similar capacity. This design avoids discrete routing decisions and preserves smooth gradient flow, which is beneficial in adversarial settings. While this may increase computational requirements compared to sparse expert selection, it reflects a deliberate trade-off favoring robustness and stability over minimal inference cost.}
\section{Experiments}
\label{sec:experiments}
In this section, we firstly describe the hardware type and software libraries used in our experiments. Then, we present the dataset, target models, and adversarial attacks considered. Next, we provide details on the normalization procedure employed during both training and test phases. We also explain the learning settings used in our proposed defense system. Finally, we introduce the evaluation metrics and demonstrate the attained results.
\begin{table*}[h]
    \centering
    \caption{Accuracy and robustness of our proposed defense system vs. other models in terms of standard accuracy (SA) and robust accuracy (RA) (FGSM) - DenseNet-169 (DN169).}
    \resizebox{\linewidth}{!}{
    \begin{tabular}{l | c | c c c c c c c c c}
        \hline
        \multirow{2}{*}{\diagbox{Model}{Metric}} & \multirow{2}{*}{SA} & \multicolumn{9}{|c}{RA - FGSM}\\

        &&0.01&0.02&0.03&0.04&0.05&0.06&0.07&0.08&0.09\\
        
        \hline
        \multicolumn{11}{>{\columncolor{green!30}}c}{CIFAR-10}\\
        \hline
        \multicolumn{11}{c}{\textbf{Publicly Available Pretrained Baselines}} \\
        \hline
        VGG-19&93.18&52.08&50.92&48.52&41.53&29.22&20.22&16.06&13.70&12.00\\
        
        ResNet-50&95.29&52.68&51.54&46.01&35.04&25.46&19.26&16.25&13.87&12.14\\
        
        DN169&95.10&55.32&54.57&48.63&36.38&25.21&18.57&15.00&12.88&11.90\\
        GoogLeNet&95.01&53.83&51.53&46.96&33.59&22.70&16.61&13.83&11.69&10.91\\
        
        Inception-V3&94.78&58.00&56.77&52.05&41.99&29.99&21.89&16.53&13.61&11.97\\
        Xception&93.48&51.41&50.20&47.69&37.72&26.35&18.45&14.21&12.32&10.96\\
        
        MobileNet-V2&94.20&55.05&53.79&49.31&36.67&24.70&18.31&15.38&13.39&12.19\\
        \hline
        \multicolumn{11}{c}{\textbf{Baselines Trained Under Our Pipeline}} \\
        \hline
        ResNet-18&92.35&52.32&48.19&40.97&30.28&21.38&15.61&12.72&11.61&11.00\\

        ResNet-18 (FGSM-AT)&89.22&78.81&67.50&48.37&32.71&23.52&16.48&12.2&9.54&9.26\\

        ResNet-18 (PGD-AT)&87.99&78.83&64.29&41.46&28.34&20.8&15.29&11.57&9.8&9.78\\
        SoE&81.69&70.23&66.22&60.27&52.65&44.62&38.26&32.96&28.74&25.74\\
        ADVMoE&88.27&79.39&62.09&40.88&23.90&16.53&15.47&15.98&16.07&16.40\\
        
        \cellcolor{blue!30}\textbf{DWF} (Ours)&91.19$\pm$0.43&\textbf{\color{blue}82.83$\pm$0.2}&\textbf{\color{blue}79.66$\pm$0.37}&\textbf{\color{blue}74.76$\pm$0.96}&\textbf{\color{blue}69.01$\pm$1.37}&\textbf{\color{blue}63.6$\pm$1.52}&\textbf{\color{blue}58.40$\pm$1.55}&\textbf{\color{blue}52.97$\pm$1.17}&\textbf{\color{blue}47.34$\pm$0.63}&\textbf{\color{blue}41.24$\pm$0.42}\\
        \hline
        \multicolumn{11}{>{\columncolor{orange!30}}c}{SVHN}\\
        \hline
        \multicolumn{11}{c}{\textbf{Publicly Available Pretrained Baselines}} \\
        \hline
        ResNet-34&96.26&70.54&67.2&63.12&57.53&48.74&38.74&28.38&19.89&14.6\\
        ResNet-50&96.32&69.02&66.75&63.03&57.98&50.81&41.39&31.33&22.6&16.62\\
        DenseNet-121&95.53&70.36&66.87&61.52&56.2&48.94&40.74&32.36&25.09&19.97\\
        \hline
        \multicolumn{11}{c}{\textbf{Baselines Trained Under Our Pipeline}} \\
        \hline
        ResNet-18&95.95&70.98&67.85&63.03&54.87&46.20&35.59&25.96&18.29&13.77\\
        ADVMoE&88.88&79.22&62.87&41.69&24.71&15.86&14.05&14.05&13.47&13.42\\
        
        \cellcolor{blue!30}\textbf{DWF} (Ours)&92.35$\pm$0.36&\textbf{\color{blue}81.43$\pm$0.22}&\textbf{\color{blue}78.02$\pm$0.42}&\textbf{\color{blue}74.88$\pm$0.99}&\textbf{\color{blue}68.94$\pm$1.59}&\textbf{\color{blue}63.1$\pm$1.36}&\textbf{\color{blue}57.53$\pm$1.68}&\textbf{\color{blue}52$\pm$1.03}&\textbf{\color{blue}47.73$\pm$0.81}&\textbf{\color{blue}41.3$\pm$0.54}\\
        \hline
        
    \end{tabular}}
    \label{tab:accuracy-robustness-all}
\end{table*}
\begin{table*}[h]
    \centering
    \caption{Accuracy and robustness of our proposed defense system vs. other models in terms of PGD robust accuracy.}
    \resizebox{\linewidth}{!}{
    \begin{tabular}{l | c c c c c c c}
        \hline
        \multirow{2}{*}{\diagbox{Model}{Metric}} & \multicolumn{7}{c}{RA - PGD}\\

        &10&20&30&40&50&100 (MS)& 100 (A)\\
        \hline
        \multicolumn{8}{>{\columncolor{green!30}}c}{CIFAR-10}\\
        \hline
        \multicolumn{7}{c}{\textbf{Publicly Available Pretrained Baselines}} \\
        \hline
        VGG-19&42.66&41.91&41.72&41.80&41.80&41.52&41.7\\

        ResNet-50&46.20&45.68&45.53&45.44&45.45&44.86&45.39\\
        
        DenseNet-169&47.55&46.99&46.81&46.72&46.81&45.72&46.73\\
        
        GoogLeNet&39.60&38.59&38.53&38.48&38.49&38.34&38.39\\
        
        Inception-V3&47.21&45.51&45.01&44.68&44.80&43.66&44.55\\
        
        Xception&41.36&40.58&40.44&40.30&40.30&40.01&40.31\\
        
        MobileNet-V2&45.77&45.20&45.14&45.17&45.08&45.03&45.12\\
        
        \hline
        \multicolumn{7}{c}{\textbf{Baselines Trained Under Our Pipeline}} \\
        \hline
        ResNet-18&37.72&37.28&37.10&37.02&36.95&37.07&37\\

        ResNet-18 (FGSM-AT)&58.65&58.21&58.17&58.14&58.13&58&58.11\\

        ResNet-18 (PGD-AT)&58.74&58.35&58.31&58.26&58.26&58.05&58.26\\
        SoE&62.27&62.34&62.31&62.31&62.31&62.03&62.27\\
        ADVMoE&47.69&44.95&44.36&44.11&44.01&42.47&43.49\\
        
        \cellcolor{blue!30}\textbf{DWF} (Ours)&\textbf{\color{blue}65.51$\pm$0.93}&\textbf{\color{blue}64.7$\pm$1.08}&\textbf{\color{blue}64.52$\pm$1.1}&\textbf{\color{blue}64.42$\pm$1.08}&\textbf{\color{blue}64.37$\pm$1.1}&\textbf{\color{blue}64.2$\pm$1.15}&\textbf{\color{blue}64.32$\pm$1.12}\\
        \hline
        \multicolumn{8}{>{\columncolor{orange!30}}c}{SVHN}\\
        \hline

        \multicolumn{7}{c}{\textbf{Publicly Available Pretrained Baselines}} \\
        \hline
        
        ResNet-34&49.04&48.04&47.83&47.71&47.66&47.37&47.54\\
        
        ResNet-50&48.74&47.88&47.7&47.6&47.54&47.25&47.45\\
        
        DenseNet-121&45.94&44.54&44.29&44.18&44.14&43.82&44.05\\
        
        \hline
        \multicolumn{7}{c}{\textbf{Baselines Trained Under Our Pipeline}} \\
        \hline
        
        ResNet-18&49.61&48.86&48.62&48.56&48.52&48.29&48.46\\
        ADVMoE&47.22&44.22&43.31&43&42.68&8.13&8.33\\
        \cellcolor{blue!30}\textbf{DWF} (Ours)&\textbf{\color{blue}64.93$\pm$1.04}&\textbf{\color{blue}64.08$\pm$1.03}&\textbf{\color{blue}63.97$\pm$1.19}&\textbf{\color{blue}63.92$\pm$1.21}&\textbf{\color{blue}63.88$\pm$1.21}&\textbf{\color{blue}63.69$\pm$1.37}&\textbf{\color{blue}63.78$\pm$1.29}\\
        \hline
        
    \end{tabular}}
    \label{tab:accuracy-robustness-all}
\end{table*}
\paragraph{Hardware and Implementation}
{All experiments were conducted on a server equipped with two NVIDIA H100 GPUs each with 80 GB of VRAM, and 512 GB of RAM, offering the adequate computational power for MoE training. While we used two NVIDIA H100 GPUs to accelerate experimentation, the proposed DWF architecture is composed of lightweight ResNet-18 experts and is modular. Training can be performed on more modest GPU environments by reducing batch size and/or using gradient accumulation; additional ablations on scaling with fewer experts are left for future work. Importantly, inference is substantially cheaper than training and can be deployed on commodity GPU hardware.} We implemented all the code in PyTorch with torch \texttt{2.7.1} and torchvision \texttt{0.22.0}, and CUDA \texttt{12.8}.

DWF introduces additional computational overhead compared to single-network baselines due to the joint training and inference of multiple experts. On our hardware, full training requires several days (approximately five days of wall-clock time, corresponding to roughly 240 GPU-hours), whereas single baselines complete substantially faster. ADVMoE and SoE exhibit intermediate training costs between these extremes. During inference, the gating module is lightweight and the dominant cost comes from evaluating the experts, resulting in approximately linear scaling with the number of experts. 

{Fractional activation evaluates all experts in parallel using continuous routing weights. As a result, inference cost scales approximately linearly with the number of experts relative to a single backbone of comparable size. The architecture does not introduce sequential routing decisions or iterative refinement steps; however, evaluating multiple experts increases FLOPs and memory proportionally to expert count. DWF is primarily a robustness-oriented design rather than a deployment-optimized solution.}
\paragraph{Datasets}
{To conduct our experiments, we use CIFAR-10 \cite{krizhevsky2009learning} and street view house number (SVHN) \cite{netzer2011reading} datasets, which are RGB-colored datasets, each include 10 labels.} 
\paragraph{Attacks}
In order to train our model and test its robustness, we employ FGSM and PGD. 
FGSM is a single step attack, which generates adversarial samples using the following formula:

\begin{equation}
\label{eq:fgsm-formula}
x_{\text{adv}} = x + \epsilon \cdot sign(\nabla_x \ell(\theta, x, y))  
\end{equation}
where $x$ is benign input, $y$ is the label, $x_{adv}$ is the perturbed version of $x$, $\epsilon$ is perturbation size, $sign$ returns the sign of the input, $\theta$ denotes the model's parameters and $\nabla_x \ell(\theta, x, y)$ is the gradient of loss function with respect to $x$.

PGD, on the other hand, is an iterative attack, computing the final version of adversarial samples iteratively:
\begin{equation}
\label{eq:PGD-Formula}
x_{adv}^{i+1} = \prod_{x + \epsilon} \left(x_{adv}^i + \alpha \cdot sign \left( \nabla_x \ell(\theta, x, y) \right) \right)
\end{equation}
where $\alpha$ is step size, $i$ is the iteration number, $x_{adv}^i$ is the perturbed sample during the $i_{th}$ iteration, and $\prod_{x + \epsilon}$ ensures each generated adversarial sample remains within $\epsilon$ perturbation bound during each iteration. The formula for $\prod_{x + \epsilon}$ is as follows:
\begin{equation}
\label{eq:projection}
x_{adv} =
\left\{
\begin{array}{l l}
     x_{adv} & \text{if} \quad x-\epsilon \leq x_{adv} \leq x+\epsilon  \\
     x+\epsilon & \text{if} \quad x_{adv} > x+\epsilon\\
     x-\epsilon & \text{if} \quad x_{adv} < x-\epsilon\\
\end{array}
\right.
\end{equation}

{Moreover, to further assess the resilience of our proposed approach, we conduct further experiments experiments under white-box multi-step PGD and adaptive PGD attacks.}

\begin{table*}[h]
    \centering   
    \caption{Attacks settings - expectation over transformation (EoT).}
    \resizebox{\linewidth}{!}{
    \begin{tabular}{l l c}
        \hline
        Attacks &  Settings & \makecell{Attacker's\\Knowledge}\\
        \hline
        FGSM & $\epsilon \in \{0.01,~0.02,...,0.09\}$&white-box\\[0.75ex]
        PGD & $\epsilon = \frac{8}{255}$, $\alpha = \frac{2}{255}$, $i\in \{10,~20,~30,~40,~50\}$&white-box\\[0.75ex]
        Multi-Step PGD&$\epsilon = \frac{8}{255}$, $\alpha = \frac{2}{255}$, $i=100$, $\text{Restart} = 3$&white-box\\
        [0.75ex]
        Adaptive PGD &$\epsilon = \frac{8}{255}$, $\alpha = \frac{2}{255}$, $i=100$, $\text{EoT Samples} = 1$&white-box\\
        \hline
    \end{tabular}}
    \label{tab:attack-settings}
\end{table*}

\begin{figure*}[h]
    \centering
    \includegraphics[width=\linewidth]{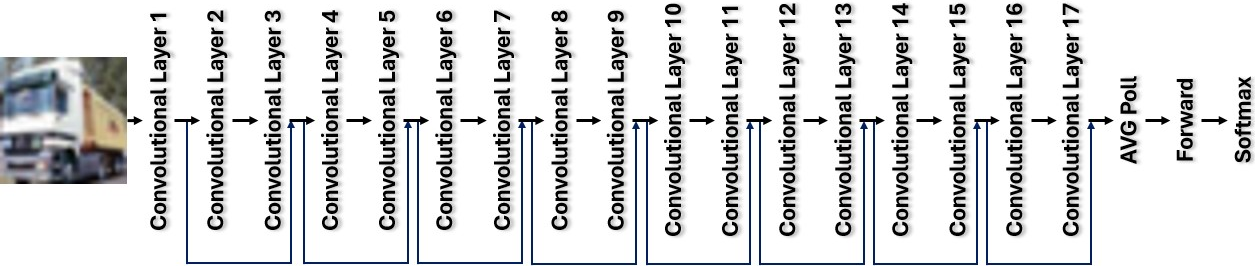}
    \caption{The backbone architecture of DWF network.}
    \label{fig:backbone}
\end{figure*}

\begin{table*}[h]
    \centering
    \caption{Hyper parameters - learning rate (LR), batch size (BS), and weight decay (WD).}
    \begin{tabular}{l c c c c}
        \hline
       Models & Epochs &  LR & BS & WD\\
       \hline
       Experts & $150, 100$ & $0.01$ & $128, 128\times 3$ & $5e-4$\\
       Gating & $100$ & $0.01$ & $128\times 3$  & $5e-4$\\
       End2End & $100$ & $0.01$ & $128\times 3$ & $5e-4$\\
       \hline
    \end{tabular}
    \label{tab:learning-setting}
\end{table*}
\paragraph{Benign baselines}
{To compare our results with plain networks we used networks pretrained on CIFAR-10 from \cite{juraev2022cifar10models} and networks pretrained on SVHN from \cite{edadaltocg2023}. These models include VGG-19 \cite{simonyan2014very}, ResNet-34, ResNet-50 \cite{he2016deep}, DenseNet-121, DenseNet-169 \cite{huang2017densely}, GoogLeNet, Inception-V3 \cite{szegedy2016rethinking}, Xception \cite{chollet2017xception}, and MobileNet-V2 \cite{sandler2018mobilenetv2}. 

Plain-network baselines are initialized from standard publicly available pretrained checkpoints. Although their original training recipes may differ slightly, all models are evaluated under an identical test-time protocol, including the same preprocessing, normalization, and attack settings, ensuring controlled robustness comparisons. These checkpoints achieve strong clean accuracy, indicating that they are competitive baselines rather than undertrained variants. The substantial robustness gap observed between our method and these baselines suggests that minor differences in the original training procedures are unlikely to affect the main conclusions.} The attained results are shown in Table \ref{tab:accuracy-robustness-all}.
\paragraph{Adversarially trained baselines}
{
To evaluate whether the observed resilience stems from the adversarial training schedule, the ensemble structure, or a combination of both, we conducted adversarial training experiments with two separate ResNet-18 models using different adversarial training methods: FGSM and PGD.}
\paragraph{Backbone}
For every expert we use ResNet-18 \cite{he2016deep} as the backbone architecture, since other relevant defenses use the same backbone. Figure \ref{fig:backbone} shows the backbone details.
\paragraph{Threat Model}
{We adopt a full white-box threat model where the adversary has complete knowledge of the entire DWF architecture, including all experts, the gating network, and routing weights. Gradients are computed through the full mixture-of-experts pipeline, allowing attacks to jointly optimize perturbations with respect to both expert predictions and gating behavior. Therefore, the attacker explicitly targets the complete MoE routing mechanism rather than individual experts only. To further approximate worst-case adversarial conditions, we additionally evaluate stronger multi-step and adaptive attacks that account for input-dependent routing, ensuring that the robustness assessment is not affected by gradient masking or routing-specific obfuscation.}

{To evaluate our model's resilience against white box attacks, we employ FGSM with the perturbation sizes $\epsilon \in \{0.01,~0.02,~\dots,~0.09\}$, and PGD with the step size $\alpha = \frac{2}{255}$, perturbation bound $\epsilon=\frac{8}{255}$, and iterations numbers $i \in \{10,~20,~30,~40,~50\}$. These setting were used during both training and test phases. Moreover, to further evaluate our proposed defense's performance against unseen attacks, we used multi-step PGD-100 with 3 restarts and adaptive PGD-100 with expectation over transformation (EoT) of one. Table \ref{tab:attack-settings} shows an overview of the attacks settings.}
\paragraph{Normalization Procedure}
Let $x \in [0,1]^{32 \times 32 \times 3}$ be an RGB image sample from CIFAR-10. The normalization procedure for each of the three channels is as follows:
\begin{equation}
\label{eq:normalization procedure}
x'_c(i,j) = \frac{x_c(i,j)-\mu_c}{\sigma_c} 
\end{equation}
where $c \in \{R,G,B\}$ stands for each color channel, $x'_c$ is the normalized version of $x_c$, $\mu_c$ is the empirical mean over channel $c$, $\sigma_c$ is the empirical deviation over channel $c$, and $i,j$ are row index and column index, respectively. To normalize {CIFAR-10} data we defined $\mu = (0.4377, ~0.4438, ~0.4728)$ and $\sigma = (0.1980, ~0.2010, ~0.1970)$. Moreover, to normalize SVHN data we defined $\mu = (0.4914,~0.4822,~0.4465)$ and $\sigma =(0.2023,~0.1994,~0.2010)$.

{For each dataset, we apply a fixed per-channel normalization to all splits (train and test) using the same preprocessing pipeline. Importantly, the same normalization parameters are used consistently across all defended models, including DWF, ADVMoE, SoE, and adversarially trained ResNet-18 models, ensuring fair and controlled comparisons.
The CIFAR-10 and SVHN preprocessing pipelines are dataset-specific and were not interchanged.}
\paragraph{Learning Settings}
We trained all the experts over $150$ epochs using batch size of $128$. We used stochastic gradient descent as the optimizer with an initial learning rate of $0.01$, momentum of $0.9$, and weight decay of $5e^{-4}$. During the training session of the end to end framework, all the models are retrained as well as the gating mechanism for $100$ epochs with batch size of $128\times 3$, where the first part includes benign inputs, the second and the third include the corresponding FGSM and PGD inputs, respectively. We also used the cross-entropy loss as the loss function for training.

Additionally, to further improve our model's stability, we employed a cosine annealing learning rate scheduler \cite{loshchilov2016sgdr} to decay the learning rate gradually over all iterations in all training sessions. Table \ref{tab:learning-setting} shows an overview of the hyper parameters used for training our proposed approach.

ADVMoE used TRADES \cite{zhang2019theoretically} as the adversarial training objective by adopting PGD with attack strength of $\epsilon=\frac{8}{255}$ during two iterations. Additionally, it was trained during 200 epochs with learning rate of 0.1, momentum of 0.9, weight decay of $5e^{-4}$, and batch size of 128. SoE was adversarially trained during 200 epochs with learning rate of 0.1 and batch size of 128. {The first ResNet-18 model was adversarially trained using the FGSM with a perturbation magnitude of $\epsilon = 0.03$, where half of each training batch consisted of FGSM-generated adversarial examples. The second ResNet-18 model was trained using PGD, with the adversarial perturbation set to $\epsilon = \frac{8}{255}$, step size $\alpha = \frac{2}{255}$, and 10 iteration, where half of each training batch consisted of PGD-generated adversarial examples.}

{To ensure fair comparisons, all methods use the same backbone architecture (ResNet-18), optimizer, learning rate schedule, and training protocol. Some baselines (e.g., ADVMoE and SoE) also employ multiple experts, although the number of experts varies across methods as part of their architectural design. A larger number of experts increases computational cost, analogous to increasing network width or depth, and should therefore be interpreted as a capacity trade-off rather than an advantage in training settings. Our comparisons keep the optimization procedure identical while allowing each method to retain its original architecture.}
\paragraph{Metrics}
\begin{figure}[h]
    \centering
    \includegraphics[width=\linewidth]{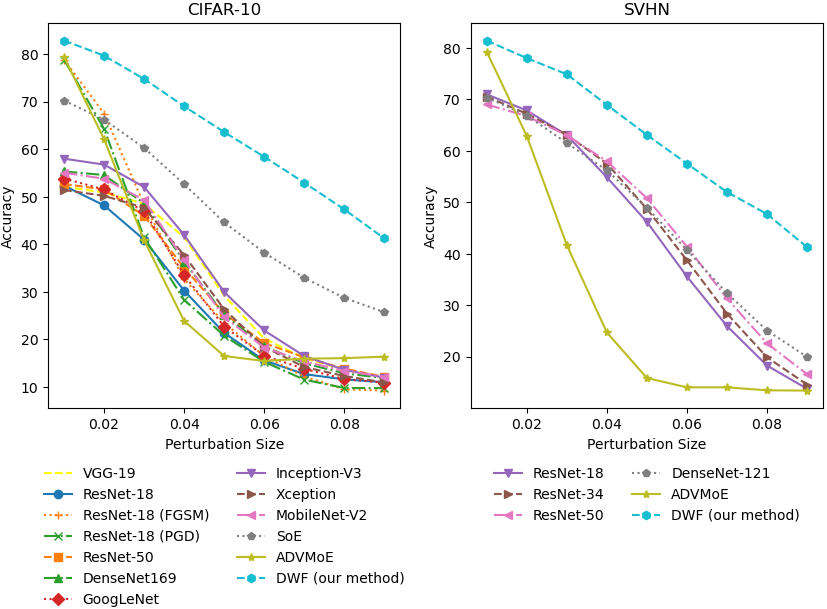}
    \caption{Robustness of our proposed model compared with the other competitors against FGSM w.r.t. different perturbation sizes.}
    \label{fig:fgsm-plot}
\end{figure}
\begin{figure}[h]
    \centering
    \includegraphics[width=\linewidth]{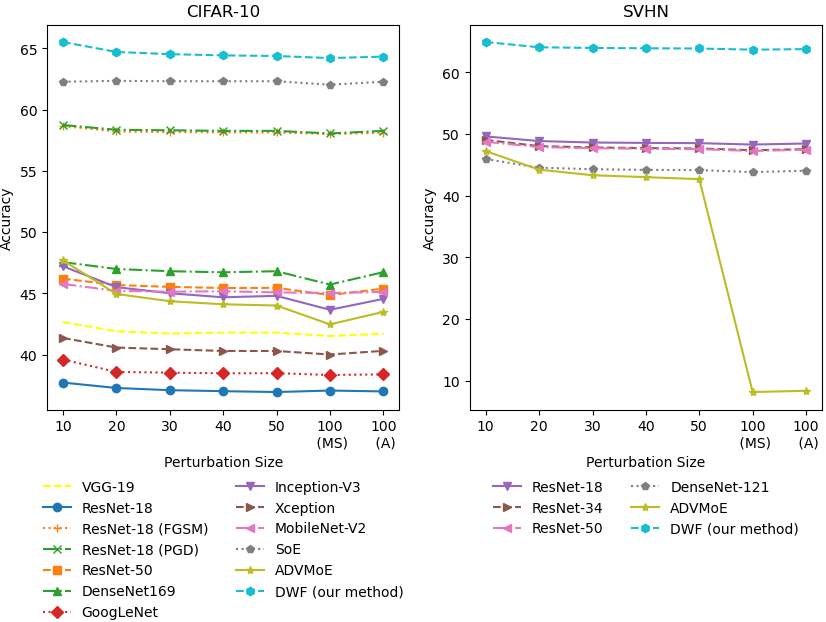}
    \caption{Robustness of our proposed model vs. other models against PGD w.r.t. different iteration numbers.}
    \label{fig:pgd-plot}
\end{figure}
To compare our model with other models, we use two metrics called standard accuracy (SA) and robust accuracy (RA). The first metric measures the model's accuracy on the benign data, while the latter one measures the models accuracy on the adversarial data.
\paragraph{Results}
Table \ref{tab:accuracy-robustness-all} illustrates the accuracy and robustness of our proposed model in comparison with other models. Figures \ref{fig:fgsm-plot} and \ref{fig:pgd-plot} present a visual summary of the results reported in Table \ref{tab:accuracy-robustness-all} and provide clearer comparison between our proposed model and other competitors in terms of robust accuracy under different settings for FGSM and PGD attacks. The attained results by our model are averaged over {five} runs, ensuring a fair comparison. {For the defended baseline methods, we follow their standard evaluation protocols and report the corresponding single-run results.}

{To improve transparency and fairness, we distinguish between publicly available pretrained baselines and models trained under our experimental pipeline. Public checkpoints are evaluated using our standardized preprocessing and attack settings but retain their original training recipes. In contrast, adversarially trained ResNet-18 baselines and MoE defenses (SoE, ADVMoE, and DWF) are trained under the same optimization protocol, preprocessing pipeline, and threat model. This separation clarifies that performance differences between MoE-based methods reflect architectural and training design choices rather than inconsistencies in evaluation settings.}

{To isolate architectural effects from training-related factors, all compared defended methods, including ADVMoE and SoE, are trained and evaluated under the same experimental pipeline with identical preprocessing, backbone configuration, and optimization protocol. Robust accuracy is computed using the same threat models for every method. Consequently, performance differences primarily reflect differences in architectural design rather than inconsistencies in training or evaluation procedures.}

The results over these runs vary slightly due to the randomness during training sessions. This randomness occurs due to four reasons: i) training data shuffling, ii) random weight initialization of the models before training, iii) random FGSM perturbation size of adversarial sample generation for each batch in every epoch, and iv) random PGD iteration of adversarial sample generation for each batch in every epoch. 

Moreover, our proposed approach suffers from {robust accuracy degradation}, when faces stronger attacks, which is very common among all other competitors. We can see that our proposed model outperforms all other competitors in terms of robustness under all attacks. {Additionally, DWF consistently outperforms existing defenses such as ADVMoE, SoE, and adversarially trained networks in terms of clean accuracy. Notably, it maintains a clean accuracy above 90\% across both datasets, highlighting its effectiveness not only against adversarial attacks but also in non-adversarial environments. Compared to plain (undefended) networks, DWF exhibits at most a modest reduction in clean accuracy, which is expected since DWF incorporates adversarial training and jointly optimizes robustness-oriented objectives through the gating mechanism and expert updates. This optimization slightly shifts capacity toward worst-case robustness, whereas plain models are optimized purely for benign performance. Importantly, the clean accuracy drop in DWF is noticeably smaller than in standard adversarially trained baselines, indicating that the mixture-of-experts design helps preserve discriminative capacity on benign inputs while still providing substantial robustness gains.} 

{Moreover, the results demonstrate that our method could preserve its performance under adaptive white-box scenario, suggesting that the resilience does not stem from gradient masking/obfuscation.} Another insight illustrates that as we increase perturbation size or iterations in FGSM and PGD, the robustness of all other models drops, which shows the lack of adaptability of the models against attacks with more extreme specifications. {As a result, it is useful to consider how the gating mechanism may behave under stronger perturbations. In a soft-gated mixture-of-experts architecture, routing can distribute weights across multiple experts when predictions become uncertain, which may help avoid complete reliance on any single expert and lead to gradual performance degradation as attack strength increases. We focus on overall robustness trends in this work.}

{The robustness gains of DWF can be interpreted through several complementary mechanisms. First, multiple experts introduce functional diversity, which reduces correlated failure modes and makes adversarial transfer across experts more difficult. Second, soft gating enables input-dependent specialization, allowing each expert to focus on a subset of the data manifold and improving local decision boundaries. Third, joint optimization with adversarial training encourages smoother and more stable gradients across the mixture. Together, these effects contribute to improved robustness without a large sacrifice in clean accuracy.}

{The proposed design choices (soft routing, fractional activation, staged pretraining, and joint optimization) are not independent modifications but structurally coupled components of a unified framework. Fractional activation inherently relies on continuous routing weights, and replacing it with sparse or hard gating alters gradient flow and stability properties. Likewise, the pretrain+joint schedule is intended to promote initial expert differentiation prior to collaborative adaptation, which differs qualitatively from joint-only training. For this reason, we evaluate the framework in its complete form, as these elements collectively define the optimization behavior of the model.}
\section{Limitations and Future Works}
\label{sec:limitations-future-works}
{In this study, we evaluated DWF's robustness primarily under white-box attacks, which represent the strongest and most conservative threat model since the adversary has full access to the model parameters and gradients. Robustness under white-box optimization is widely regarded as a lower bound on performance under weaker settings, such as transfer-based or query-limited black-box attacks, which typically yield higher robust accuracy. Therefore, defenses that remain effective in the white-box scenario are expected to generalize well to these weaker threat models. A systematic evaluation of DWF under transfer and black-box attacks is an interesting direction for future work.

DWF integrates several complementary mechanisms, including soft gating, fractional expert activation, joint expert-gating optimization, and randomized adversarial training. Each component addresses a distinct limitation of conventional defenses. Soft gating enables input-dependent expert specialization, fractional activation reduces overfitting and improves diversity, joint updating ensures coordinated optimization between experts and routing, and randomized adversarial training improves robustness across varying perturbation strengths. Together, these components contribute synergistically to both clean and robust accuracy. Conducting a full combinatorial ablation over all variants would require retraining numerous MoE configurations with substantial computational cost; therefore, we leave a comprehensive ablation study for future work.

While this work focuses primarily on improving robustness, training efficiency is also an important practical consideration. Fast adversarial training variants (e.g., FFAT or single-step approximations) may reduce computational cost at the expense of some robustness. A systematic study quantifying the trade-off between training time and robust accuracy for DWF under different efficient training strategies would require multiple full training runs and extensive hyperparameter tuning. We leave this investigation for future work.

Although the proposed framework is evaluated on image classification tasks, the soft-gated mixture-of-experts design is largely architecture-agnostic and can, in principle, be integrated with other modalities such as text, speech, or time-series models. The core idea of training specialized experts and adaptively routing inputs through a gating network does not depend on convolutional backbones. However, extending the method to other domains may require modality-specific adversarial threat models and training strategies, and could introduce additional computational costs. Exploring these extensions is an interesting direction for future work.

In summary, the main practical constraints of DWF stem from computational cost, scaling behavior, and evaluation scope. Training multiple experts jointly increases training time and memory usage compared to single-network defenses, and inference cost scales approximately linearly with the number of experts, which may limit deployment in resource-constrained environments. While our evaluation focuses on strong white-box evasion attacks, robustness under broader threat models (such as transfer-based, query-limited black-box, poisoning, or physical attacks) remains to be systematically assessed.}
\section{Conclusion}
\label{sec:conclusion}
{In this paper, we introduced DWF, a soft-gated fractional mixture-of-experts defense that combines joint expert-router optimization with randomized adversarial training to improve robustness under a white-box $\ell_\infty$ threat model. Extensive experiments on CIFAR-10 and SVHN demonstrate that DWF consistently achieves higher robust accuracy than prior MoE-based defenses (SoE and ADVMoE) and adversarially trained single-network baselines across FGSM, multi-step PGD, and adaptive PGD attacks. Notably, these robustness gains are obtained while preserving high clean accuracy (above 90\%), indicating that the proposed mixture-of-experts design mitigates the traditional robustness-accuracy trade-off.

The results show that soft gating with fractional expert activation, combined with a pretrain-then-joint training pipeline and randomized adversarial training, enables cooperative expert behavior and more stable predictions under adversarial perturbations. By isolating routing, activation, and training design within a unified MoE framework, DWF demonstrates that improving expert utilization and training dynamics can yield substantial robustness gains without requiring major architectural changes. These findings highlight the practical potential of soft-gated MoE defenses for strengthening adversarial resilience in image classification systems.}
\bibliographystyle{ieeetr}
\bibliography{main}

\end{document}